\documentclass[conference]{IEEEtran}
\ifCLASSINFOpdf
\else
\fi
\usepackage{graphicx}
\usepackage{multirow}
\usepackage{comment}
\usepackage[noadjust]{cite}

\usepackage{algorithm}
\usepackage{algorithmic}

\usepackage{times}
\usepackage{epsfig}
\usepackage{amsmath}
\usepackage{amssymb}
\usepackage{booktabs}
\usepackage{array}
\usepackage{multirow}

\begin{document}
%
\title{Leveraging progressive model and overfitting for efficient learned
image compression}

\author{
\IEEEauthorblockN{Honglei Zhang, 
  Francesco Cricri,
  Hamed Rezazadegan Tavakoli,
  Emre Aksu,
  Miska M. Hannuksela
}
\IEEEauthorblockA{
Nokia Technologies\\
Finland}

}


%


\maketitle

\begin{abstract}

Deep learning is overwhelmingly dominant in the field of computer
vision and image/video processing for the last decade. However, for
image and video compression, it lags behind the traditional techniques
based on discrete cosine transform (DCT) and linear filters. Built
on top of an auto-encoder architecture, learned image compression
(LIC) systems have drawn enormous attention in recent years. Nevertheless,
the proposed LIC systems are still inferior to the state-of-the-art
traditional techniques, for example, the Versatile Video Coding (VVC/H.266)
standard, due to either their compression performance or decoding
complexity. Although claimed to outperform the VVC/H.266 on a limited
bit rate range, some proposed LIC systems take over 40 seconds to
decode a 2K image on a GPU system. In this paper, we introduce a powerful
and flexible LIC framework with multi-scale progressive (MSP) probability
model and latent representation overfitting (LOF) technique. With
different predefined profiles, the proposed framework can achieve
various balance points between compression efficiency and computational
complexity. Experiments show that the proposed framework achieves
2.5\%, 1.0\%, and 1.3\% Bjontegaard delta bit rate (BD-rate) reduction
over the VVC/H.266 standard on three benchmark datasets on a wide
bit rate range. More importantly, the decoding complexity is reduced
from $O(n)$ to $O(1)$ compared to many other LIC systems, resulting
in over 20 times speedup when decoding 2K images.

\end{abstract}


%
\IEEEpeerreviewmaketitle


\section{Introduction}

\begin{figure}
\begin{centering}
\includegraphics[width=8cm]{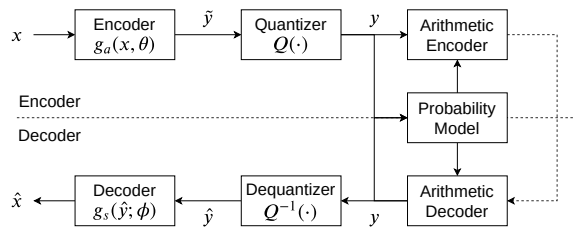} 
\par\end{centering}
\caption{LIC architecture. \label{fig:LIC architecture} }
\end{figure}

Although deep learning-based technology has achieved tremendous success
in most computer vision and image/video processing tasks, it has not
been able to demonstrate superior performance over traditional technologies
for image and video compression, in particular, for practical usage.
Traditional image compression techniques, such as JPEG/JPEG 2000 \cite{jpeg},
High Efficiency Video Coding (HEVC) (all-intra mode) \cite{sullivan2012overview},
and Versatile Video Coding (VVC/H.266) (all-intra mode) \cite{standardization2021isoiec2309032021},
apply carefully designed processing steps such as data transformation,
quantization, entropy coding to compress the image data while maintaining
certain quality for human perception \cite{hu2021learning,marpe2006theh264mpeg4,tan2015videoquality}.
End-to-end learned image compression (LIC) systems are based on deep
learning technology and data-driven paradigm \cite{balle2017endtoend,cheng2020learned,guo2021causalcontextual,hu2020coarsetofine,minnen2018jointautoregressive,theis2017lossyimage,choi2019variable,patel2019humanperceptual,mentzer2019conditional2,hu2021learning,zhao2021auniversal2,he2021checkerboard}.
These systems normally adopt the variational auto-encoder architecture,
as shown in Figure \ref{fig:LIC architecture}, comprising encoder,
decoder and probability model implemented by deep convolutional neural
networks (CNN) \cite{balle2017endtoend,balle2018variational,cheng2020learned,guo2021causalcontextual,hu2020coarsetofine,minnen2018jointautoregressive,theis2017lossyimage,choi2019variable,patel2019humanperceptual}.
LIC systems are trained on datasets with a large number of natural
images by optimizing a rate-distortion (RD) loss function.

Compared with the state-of-the-art traditional image compression technologies,
for example, VVC/H.266, most LIC systems do not provide better compression
performance despite a much higher encoding and decoding complexity
\cite{begaint2020compressai,hu2021learning,patel2019humanperceptual}.
Recently, some proposed LIC systems have improved the compression
performance to be on par with or slightly better than the VVC/H.266
\cite{fu2021learned,gao2021neuralimage,guo2021causalcontextual,ma2021across,xie2021enhanced,he2021checkerboard2}
on a limited bit rate range. However, the decoding procedures of these
systems are very inefficient which prevents them from being used in
practice.

In this paper, we propose a flexible and novel LIC framework that
achieves various balance points between compression efficiency and
computational complexity. A system based on the proposed frame outperforms
the VVC/H.266 on three benchmarking datasets over a wide bit rate
range. Compared to most other LIC systems, the proposed system not
only improves the compression performance but also reduces the decoding
complexity from $O(n)$ to $O(1)$ in a parallel computing environment.
Our contributions are summarized as follows: 
\begin{itemize}
\item We propose the multi-scale progressive (MSP) probability model for
lossy image compression that efficiently exploits both spatial and
channel correlation of the latent representation and significantly
reduces decoding complexity. 
\item We present a greedy search method in applying the latent representation
overfitting (LOF) technique and show that LOF can considerably improve
the performance of LIC systems and mitigate the domain-shift problem.
\end{itemize}

\section{LIC system and related works\label{sec:LIC-system-and}}

In \cite{balle2017endtoend}, the authors formulated the LIC codec,
named as transform coding model, from the generative Bayesian model.
Figure \ref{fig:LIC architecture} shows the architecture of a typical
LIC codec. Input data $x$ is transformed by an analysis function
$g_{a}\left(x;\theta\right)$ to generate a latent representation
$\tilde{y}$ in continuous domain. Next, $\tilde{y}$ is quantized
to latent representation $y$ in discrete domain. Then, the arithmetic
encoder encodes $y$ into a bitstream using the estimated distribution
provided by the probability model. The encoder operation can be represented
by the function

\begin{equation}
y=Q\left(g_{a}\left(x;\theta\right)\right).\label{eq:latent_derivation}
\end{equation}
At the decoder side, an arithmetic decoder reconstructs $y$ from
the bitstream with the help of the same probability model that is
used at the encoder side. Next, $y$ is dequantized and a synthesis
function $g_{s}\left(y;\phi\right)$ is used to generate $\hat{x}$
as the reconstructed input data. $g_{a}(\cdot;\theta)$ and $g_{s}\left(\cdot;\phi\right)$
are implemented using deep neural networks with parameters $\theta$
and $\phi$, respectively. The codec is trained by optimizing the
RD loss function defined by 
\begin{alignat}{1}
L & =R\left(y\right)+\lambda\cdot D\left(x,\hat{x}\right)\label{eq:RD_loss_overall}\\
 & =\mathbb{E}_{x\sim p(x)}\left[-\log p\left(y\right)\right]+\lambda\cdot\mathbb{E}_{x\sim p(x)}\left[d\left(x,\hat{x}\right)\right]\label{eq:RD_loss}
\end{alignat}
In Eq. \ref{eq:RD_loss_overall}, $R(y)$ is the rate loss measuring
the expected number of bits to encode $y$, $D\left(x,\hat{x}\right)$
is the expected reconstruction loss measuring the quality of the reconstructed
image, and $\lambda$ is the Lagrange multiplier that adjusts the
weighting of the two loss terms to achieve difference compression
rate. In Eq. \ref{eq:RD_loss}, $p\left(y\right)$, also known as
prior distribution, is the probability distribution of latent representation
$y$, $d\left(x,\hat{x}\right)$ is the distance function measuring
the quality of the reconstructed data $\hat{x}$, where MSE or MS-SSIM
is normally used as the reconstruction loss \cite{hu2021learning,wang2003multiscale}.
The RD loss function is a weighted sum of the expected bitstream length
and the reconstructed loss.

The rate loss $R\left(y\right)$ is calculated by the expected length
of the bitstream to encode latent representation $y$ when input data
$x$ is compressed. Let $q\left(y\right)$ be the true distribution
of $y$. Although $y$ is deterministic over random variable $x$
according to Eq. \ref{eq:latent_derivation}, $q\left(y\right)$ is
still unknown since the true distribution of $x$ is unknown. To tackle
this problem, an LIC codec uses a variational distribution $p\left(y\right)$,
either a parametric model from a known distribution family \cite{balle2018variational,chen2019neuralimage,cheng2020learned,fu2021learned,guo2021causalcontextual,ma2021across,minnen2018jointautoregressive,minnen2020channelwise}
or a non-parametric model \cite{balle2017endtoend}, to replace $q\left(y\right)$
in calculating the rate loss. The rate loss is then calculated as
the cross-entropy between $q\left(y\right)$ and $p\left(y\right)$,
such as

\begin{alignat}{1}
R\left(y\right) & =H\left(q\left(y\right),p\left(y\right)\right)\label{eq: cross_entropy_loss}\\
 & =\mathbb{E}_{y\sim q\left(y\right)}\left[-\log p\left(y\right)\right]\\
 & =H\left(q\left(y\right)\right)+D_{KL}\left(q\left(y\right)\parallel p\left(y\right)\right).\label{eq:rate_loss_KL}
\end{alignat}

In \cite{balle2017endtoend}, the authors model the prior distribution
$p\left(y\right)$ as fully factorized where the distribution of each
element is modeled non-parametrically using piece-wise linear functions.
This simple model does not capture spatial dependencies in the latent
representation. In \cite{balle2018variational}, the authors introduced
the scale hyperprior model, where the elements in the latent representation
are modeled as independent zero-mean Gaussian distributions and the
variance of each element is derived from side information $z$. $z$
is modeled using a different distribution model and transferred separately.
The loss function becomes 
\begin{equation}
L=R\left(y\right)+R\left(z\right)+\lambda D\left(x,\hat{x}\right).\label{eq:Hyperprior_RD_loss}
\end{equation}

In \cite{minnen2018jointautoregressive}, the authors improved the
scale hyperprior model from two aspects. First, both the mean and
the variance of the Gaussian model are derived from the side information
$z$, named as mean-scale hyperprior model. Second, a context model
is introduced to further exploit the spatial dependencies of the elements.
The context model uses the elements that have already been decoded
to improve the model accuracy of the current element. A similar technique
was used in image generative models such as PixelCNN \cite{Jain_Abbeel_Pathak_2020,oord2016conditional,salimans2016pixelcnn}.
Many recent LIC systems are based on the hyperprior architecture and
the context model. The authors in \cite{cheng2020learned} use mixture
of Gaussian distributions instead of the Gaussian distribution in
the mean-scale hyperprior model. In \cite{fu2021learned}, the authors
further apply mixture of Gaussian-Lapalacian-Logistic (GLL) distribution
to model the latent representation. In \cite{chen2019neuralimage},
the authors enhance the context model by exploiting the channel dependencies.
The parameters of the distribution function of the elements in $y$
are derived from the channels that have already been decoded. A 3D
masked CNN is used to improve computational throughput. In \cite{guo2021causalcontextual},
the authors divide the channels into two groups, where the first group
is decoded in the same way as the normal context model and the second
group is decoded using the first group as its context. With this architecture,
the pixels in the second group are able to use the long-range correlation
in $y$ since the first group is fully decoded. In \cite{he2021checkerboard},
the authors proposed a method that partitions the elements in the
latent representation into two groups along the spatial dimension
in a checkerboard pattern. The elements in the first group are used
as the context for the elements in the second group. 

The context model, inspired by PixelCNN, exploits the spatial and
channel correlation further. Although the encoding can be performed
in a batch mode, the main issue of the PixelCNN-based context model
is that the decoding has to be performed in sequential order, i.e.,
pixel by pixel, or even element by element if the channel dependency
is exploited. According to the evaluation reported in \cite{begaint2020compressai},
in an environment with GPUs, the average decoding time of Cheng2020
model \cite{cheng2020learned} is 5.9 seconds per image with a resolution
of $512\times768$ and 45.9 seconds per image with an average resolution
of $1913\times1361$. To achieve better compression performance, recent
LIC systems are even multiple times slower than the Cheng2020 system
\cite{fu2021learned,guo2021causalcontextual,ma2021across}. Furthermore,
to avoid excessive computational complexity, the context model is
implemented with a small neural network with a limited receptive field,
which greatly degrades the system performance. The hyperprior architecture
also tries to capture the spatial correlation in the latent representation.
However. this architecture significantly increases the system complexity,
for example, a 4-layer context model network and a 5-layer hyperpior
decoder network are used in \cite{cheng2020learned}. 

In this paper, we propose a LIC framework using a novel probability
model which significantly improves the compression performance and
decoding efficiency. Table \ref{tab:related_work} summarizes previously
proposed LIC systems and the proposed system in this paper. In the
table, full channel dependency mode means that each channel is dependent
on all previous channels. Group dependency mode means that the channels
are divided into groups and the channel dependency modes are different
in each group or between groups. The decoding complexity measures
the complexity of decoding operation in a parallel computing environment.
$n$ is the number of pixels in the latent representation, $c$ is
the number of channels. $O\left(1\right)$ means all pixels can be
processed in parallel in a constant number of steps.

\begin{table*}
\caption{Comparison of the proposed method to prior works, in terms of characteristics
and complexity of the probability model.\label{tab:related_work}}

\smallskip{}

\begin{centering}
\begin{tabular}{l>{\centering}p{3cm}>{\centering}m{3cm}>{\centering}m{2cm}>{\centering}m{2cm}>{\centering}m{2cm}}
\toprule 
method  & Hyperpior architecture  & Distribution model  & Context Model  & Channel Dependency  & Decoding complexity\tabularnewline
\midrule 
Balle2017 \cite{balle2017endtoend}  & No  & non-parametric  & No  & None  & $O\left(1\right)$\tabularnewline
Balle2018 \cite{balle2018variational}  & Yes  & $\mathcal{N}\left(0,\sigma^{2}\right)$  & No  & None  & $O\left(1\right)$ \tabularnewline
Minnen2018 \cite{minnen2018jointautoregressive}  & Yes  & $\mathcal{N}\left(\mu,\sigma^{2}\right)$  & PixelCNN  & None  & $O\left(n\right)$\tabularnewline
Chen2019 \cite{chen2019neuralimage}  & Yes  & $\mathcal{N}\left(\mu,\sigma^{2}\right)$  & PixelCNN  & Full  & $O\left(cn\right)$\tabularnewline
Cheng2020 \cite{cheng2020learned}  & Yes  & Gaussian mixture  & PixelCNN  & None  & $O\left(n\right)$\tabularnewline
He2021 \cite{he2021checkerboard} & Yes & Gaussian mixture & Checkerboard & None & $O\left(1\right)$ \tabularnewline
Fu2021 \cite{fu2021learned}  & Yes  & GLL mixture  & PixelCNN  & None  & $O\left(n\right)$\tabularnewline
Ma2021 \cite{ma2021across}  & Yes  & $\mathcal{N}\left(\mu,\sigma^{2}\right)$  & PixelCNN  & Group  & $O\left(kn\right)$\dag{}\tabularnewline
Guo2021 \cite{guo2021causalcontextual}  & Yes  & Gaussian mixture  & PixelCNN  & Full  & $O\left(cn\right)$ \tabularnewline
\midrule 
Proposed  & No  & $\mathcal{N}\left(\mu,\sigma^{2}\right)$  & MSP  & Group  & $O\left(1\right)$ \ddag{}\tabularnewline
\bottomrule
\end{tabular}
\par\end{centering}
\smallskip{}

\dag : $k=c/g+g-1$, where $g$ is the number of groups. Strictly
speaking, this method is at least $O\left(\sqrt{c}n\right)$.

\ddag : The decoding takes $k$ steps, where $k$ is a predefined
constant number. The maximum $k$ we used in our experiments is 121. 
\end{table*}

\section{Methods}

According to Eq. \ref{eq:rate_loss_KL}, the compression performance
is determined by two terms: the entropy of latent representation $y$
and the KL divergence between the true distribution $q\left(y\right)$
and the variational distribution $p\left(y\right)$. If the analysis
transform function $g_{a}\left(\cdot;\theta\right)$ is powerful enough
to transform the true distribution $q\left(y\right)$ to the variational
distribution $p\left(y\right)$, the KL divergence term is minimized.

In \cite{balle2018variational}, the authors noticed that the divergence
between the two distributions is quite significant. Clear correlations
are observed among the pixels in the latent representation when the
elements in $y$ is modeled to be mutually independent, i.e., $p\left(y\right)=\prod_{i}^{cn}p\left(y_{i}\right)$.
To capture the correlation among the pixels, the hyperprior approach
is proposed, where the pixels in $y$ is modeled to be mutually independent
on the condition of a side information $z$, i.e. $p\left(y\right)=\prod_{i}^{cn}p\left(y_{i}|z\right)$.
The side information $z$, derived from transforming $y$ using a
hyperprior analysis function $g_{h}(\cdot)$, is transferred in addition
to $y$ using another variational distribution $p\left(z\right)$
with the assumption that the elements are mutually independent, i.e.,
$p\left(z\right)=\prod_{i}^{c_{z}n_{z}}p\left(z_{i}\right)$, where
$c_{z}$ and $n_{z}$ are the number of channels and the number of
pixels in $z$, respectively. With the help of the side information
$z$, correlation of the elements in $z$ can be expressed in $p\left(y\right)$.
However, the size of $z$ has to be small, otherwise, the compression
performance is impaired by the KL divergence of the true distribution
$q\left(z\right)$ and the variational distribution $p\left(z\right)$.
Furthermore, the hyperprior branch increases the complexity of the
system and makes it more difficult to train. We argue that a properly
defined variational distribution $p\left(y\right)$ is more effective
than using side information to reduce the KL divergence and improve
the compression efficiency.

\subsection{Spatial dependencies}

In \cite{minnen2018jointautoregressive}, a context model inspired
by the autoregressive image generation model PixelCNN \cite{oord2016conditional,salimans2016pixelcnn}
was introduced. The pixels are processed in a raster scan order and
each pixel is dependent on the pixels above and to the left of it.
By using the joint distribution model designed for natural images
in latent space, the compression performance can be significantly
improved. As stated in Section \ref{sec:LIC-system-and}, the pixels
can only be decoded in a sequential order which makes the decoding
very inefficient.

In \cite{cao2020lossless,zhang2020lossless}, the authors proposed
a multi-scale progressive (MSP) probability model for lossless image
compression. The MSP model is based on a factorized form of the distribution
function for natural images and significantly improves the performance
of lossless image compression. In this paper, we adopt the principle
of the MSP probability model in our LIC system to replace the inefficient
PixelCNN-based context model, and avoid the use of hyperprior. Unlike
the autoregressive model used in the PixelCNN where each pixel in
image $x$ is dependent on all previous pixels, the MSP model divides
pixels into groups. The pixels in one group are dependent on the pixels
in all previous groups but mutually independent of each other in the
same group. Next, we show how the pixels in $y$ are grouped and the
factorized form of $p\left(y\right)$.

We first downsample the latent representation $y$ along the spatial
dimensions $s$ times, thus obtaining $s$ scales. For simplicity,
we use index 0 to represent the original resolution. Let $y^{(i)}$,
$i=0,1,\cdots,s$, be the set of the pixels in the $i$-th scale of
the latent representation excluding the pixels in $y^{(i+1)},y^{(i+2)},\cdots,y^{(s)}$.
Note that $y=\bigcup_{i=0}^{s}y^{(i)}$.

Let $C^{(i)}=y^{(i+1)}\cup\cdots\cup y^{(s)}$ be the context for
$y^{(i)}$. Using the autoregressive model for the groups, we have
\begin{equation}
p\left(y\right)=\prod_{i=0}^{s-1}p\left(y^{(i)}\vert C^{(i)}\right)p\left(y^{(s)}\right).\label{eq:group_dependency}
\end{equation}

\begin{figure}
\begin{centering}
\includegraphics[width=6cm]{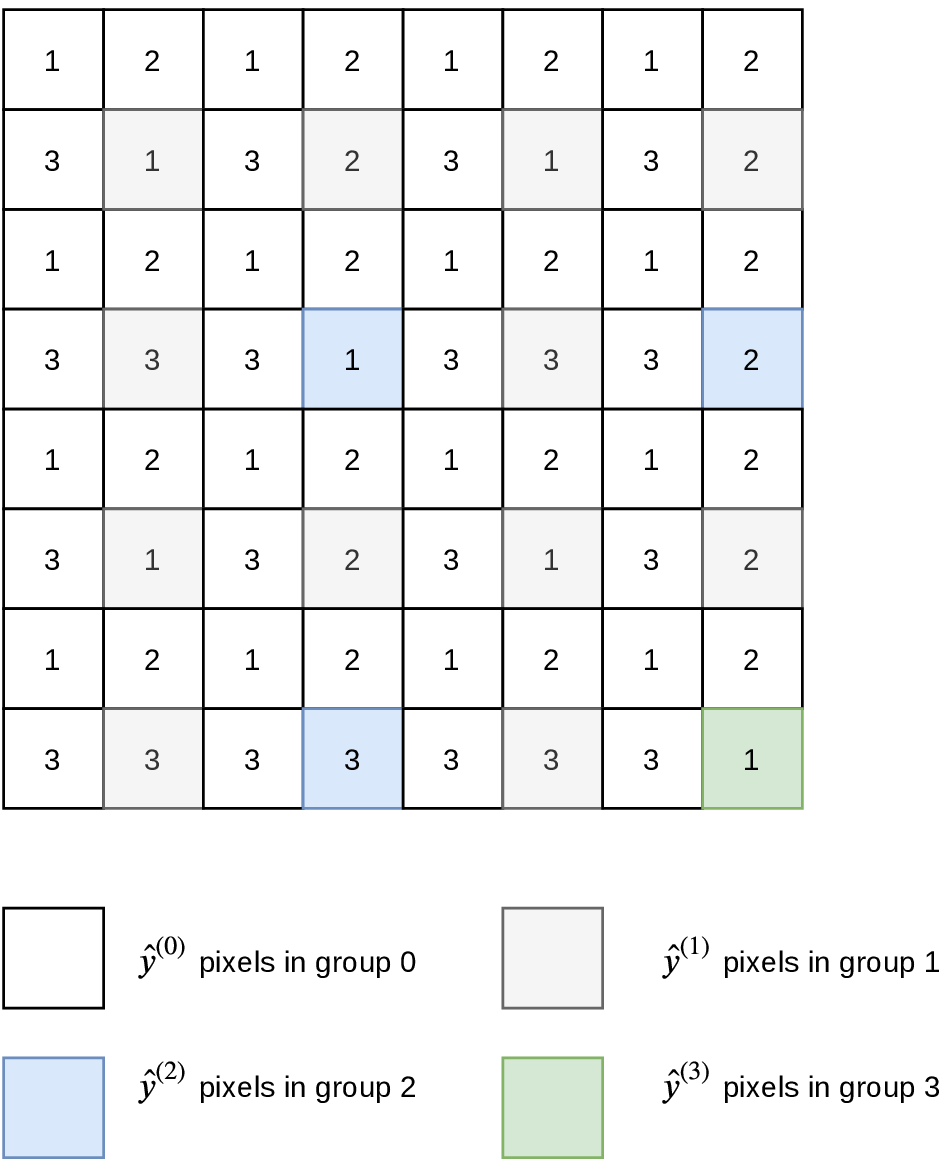} 
\par\end{centering}
\smallskip{}

\caption{Pixel groups and subgroups when an $8\times8$ latent representation
$y$ is downsampled 3 times by a factor of 2. The color indicates
the division of groups and the number in each block indicates the
subgroup index that the pixel belongs to. $2\times2$ subgroup blocks
are marked with thicker borders. \label{fig:pixel_dependencies} }
\end{figure}

To further improve the distribution model, we divide the pixels in
group $y^{(i)}$ into $b$ subgroups and model the subgroups using
the autoregressive model. Let $y^{(i,j)}$ be the $j$-th subgroup
in group $y^{(i)}$, where $i=0,\cdots,s-1$. Let $C^{(i,j)}=y^{(i,j-1)}\cup\cdots\cup y^{(i,1)}\cup C^{(i)}$
be the context of $y^{(i,j)}$. We have 
\begin{equation}
p\left(y^{(i)}\vert C^{(i)}\right)=\prod_{j=1}^{b}p\left(y^{(i,j)}\vert C^{(i,j)}\right).\label{eq:subgroup_dependency}
\end{equation}
To divide the pixels into subgroups, we first partition $y^{(i)}$
into blocks, named subgroup blocks. Each pixel is assigned to a subgroup
according to its position in the subgroup block. Figure \ref{fig:pixel_dependencies}
shows the groups and subgroups of the pixels when an $8\times8$ latent
representation $y$ is downsampled 3 times by a factor of 2 and with
the subgroup blocks of size $2\times2$.

Next, we assume that the pixels in each subgroup are mutually independent
on the condition of their context. Let $y_{(k)}^{(i,j)}$ be the pixel
$k$ in $y^{(i,j)}$ and $d$ be the total number of pixels in $y^{(i,j)}$.
We have 
\begin{equation}
p\left(y^{(i,j)}\vert C^{(i,j)}\right)=\prod_{k=1}^{d}p\left(y_{(k)}^{(i,j)}\vert C^{(i,j)}\right).\label{eq:pixel_dependency}
\end{equation}

Eqs. \ref{eq:group_dependency}, \ref{eq:subgroup_dependency} and
\ref{eq:pixel_dependency} defines a factored form of the variational
distribution function $p\left(y\right)$. It can be seen that the
PixelCNN is a special case of the MSP model when $s=0$, $b=n$ and
subgroups are designed to be in a raster scan order.

In theory, we can downsample $y$ multiple times until there is 1
pixel in the last scale. \cite{zhang2020lossless} shows that there
is only 0.4\% of data at scale 4. Thus choosing $s$ to be $3$ or
$4$ is enough in practice.

\subsection{Channel dependencies \label{subsec:Channel-dependencies}}

Channel dependency is another important aspect to build a powerful
variational distribution function \cite{chen2019neuralimage,guo2021causalcontextual,ma2021across,minnen2020channelwise}.
In \cite{zhang2020lossless}, a full channel dependency mode is used
in lossless image compression. However, when the number of channels
increases, the complexity of the full channel model increases exponentially,
which prevents it from being used in LIC systems. Next, we analyze
the properties of the channels in the latent representation. 

\begin{figure}
\begin{centering}
\setlength{\tabcolsep}{1pt}%
\begin{tabular}{cc}
\includegraphics[height=3.6cm]{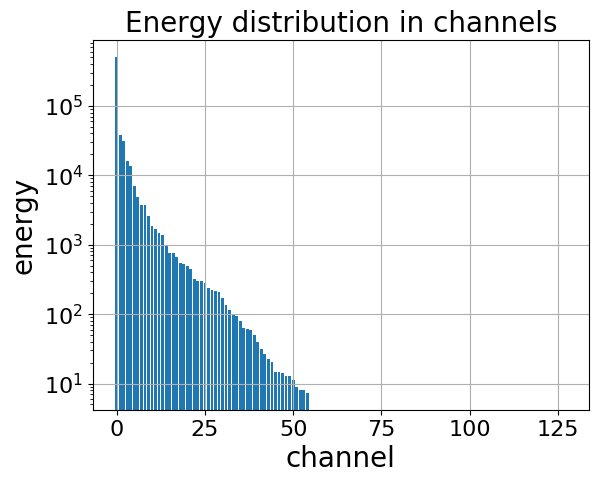}  & \includegraphics[height=3.6cm]{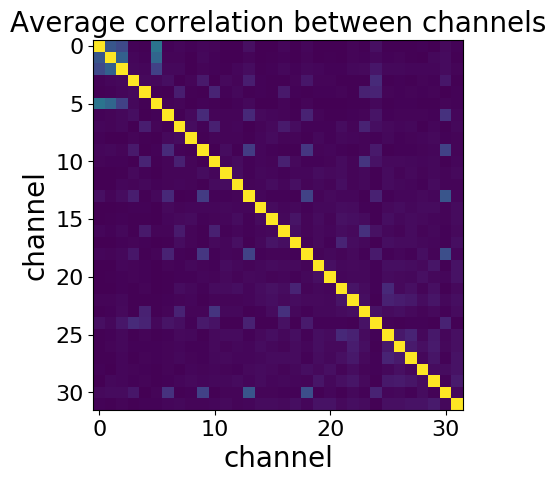}\tabularnewline
(a)  & (b)\tabularnewline
\end{tabular}
\par\end{centering}
\smallskip{}

\caption{(a) Energy distribution of the channels in $y$. (b) Correlation between
the first 32 channels sorted by their energy. \label{fig:channel_study} }
\end{figure}

First, the energy distribution is heavily unbalanced in the channels.
Figure \ref{fig:channel_study} (a) shows the energy distribution
of channels in $y$ on data collected from compressing Kodak dataset
\cite{truecolor} using a LIC system without channel dependency exploitation.
The first 32 channels account for 99.9\% of the overall energy of
the total 128 channels. The low energy channels are very sparse, for
example, only 0.1\% of the elements are non-zeros in the last 96 channels. 

Next, unlike in the image domain where the channels are highly correlated,
the correlation among the channels is low. Figure \ref{fig:channel_study}
(b) shows the average correlation between the top 32 channels. Only
a few pairs of channels show a meaningful level of correlation among
all the 32 channels.

\begin{figure*}
\begin{centering}
\includegraphics[width=14cm]{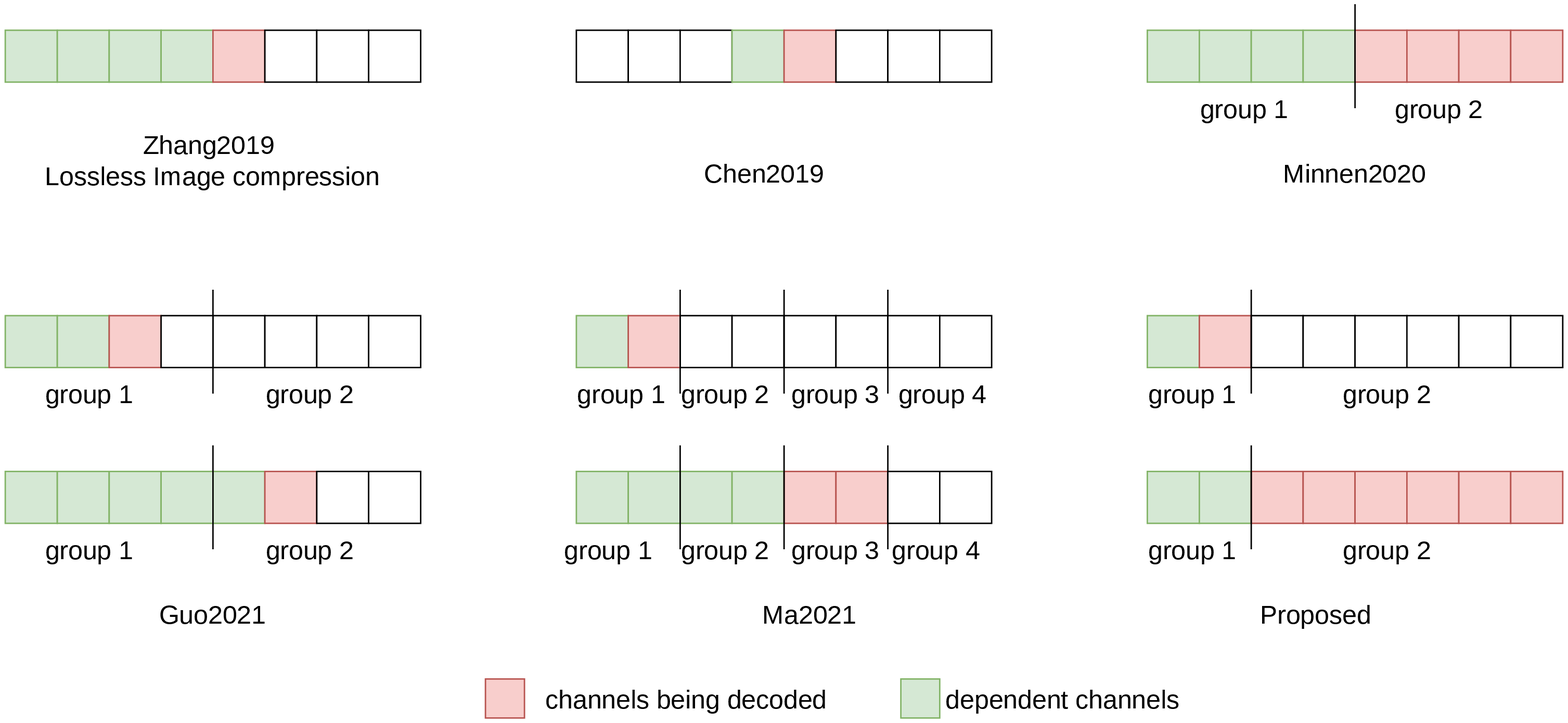} 
\par\end{centering}
\smallskip{}

\caption{Channel dependency modes of different LIC systems \label{fig:channel_deps} }
\end{figure*}

Based on these observations, exploiting channel dependencies among
a large number of channels is not necessary. we partition the channels
into two groups. The first group, named as seed group, contains $a$
number of channels and the second group contains the rest of the channels.
The channels in the seed group are processed in the full dependency
mode while the channels in the second group are processed in parallel
with dependency to the seed group. The full dependency mode for the
seed group does not increase the complexity dramatically since $a$
is a small number. In our experiments, up to $4$ channels in the
seed group are enough to capture the dependencies among the channels.
Figure \ref{fig:channel_deps} shows the channel dependency modes
of the proposed system and other LIC systems.

The probability distribution of pixel $y_{(k)}$ is

\begin{equation}
\begin{aligned} & p\left(y_{(k)}\right)=\prod_{l=2}^{a}p\left(y_{(k,l)}\vert y_{(k,l-1)},\cdots,y_{(k,1)},C\right)\\
 & \quad p\left(y_{(k,1)}\vert C\right)\prod_{l=a+1}^{c}p\left(y_{(k,l)}\vert y_{(k,a)},\cdots y_{(k,1)},C\right)
\end{aligned}
\label{eq:chnl_factor}
\end{equation}
where $y_{(k,l)}$ is the $l$-th channel element in pixel $k$, $c$
is the number of channels and $C$ is the context of pixel $k$. Note
that, superscript $(i,j)$ is omitted in Eq. \ref{eq:chnl_factor}
for simplicity.

\subsection{MSP probability model architecture}

As defined in Eqs. \ref{eq:group_dependency}, \ref{eq:subgroup_dependency},
\ref{eq:pixel_dependency} and \ref{eq:chnl_factor}, we model the
variational distribution $p\left(y\right)$ in a factorized form where
the dependencies among the elements are sufficiently exploited. The
elements in $y$ are divided into groups, subgroups, and channel groups.
The elements in the same channel group are mutually independent on
the condition of their context and are processed in parallel. The
context of each channel group contains all the elements that are available.

At last, we assume that each element in $y$ follows a Gaussian distribution
with parameter $\mu$ and $\sigma^{2}$ for the mean and variance,
respectively. $\mu$ and $\sigma^{2}$ are determined by function
$h_{p}(C)$, where $C$ is the context for the element. The cross-entropy
loss defined in Eq. \ref{eq: cross_entropy_loss} is calculated by
Eqs. \ref{eq:group_dependency}, \ref{eq:subgroup_dependency}, \ref{eq:pixel_dependency}
and \ref{eq:chnl_factor} using the derived parameters for the Gaussian
distribution functions.

Given the factorized form of the variational distribution function,
the decoder takes $s\cdot b\cdot(a+1)+1$ steps to decode the latent
representation from the bitstream. Since $s$, $b$ and $a$ are predefined
constants, the decoding complexity of the proposed LIC system is $O(1)$.

To implement the grouped dependencies of the elements in $y$, we
introduce a mixture tensor $\bar{y}$, as context $C$, that contains
the true values of $y$ that are already available and the predicted
values of $y$ that have not been available at the decoding stage.
A binary tensor $m$ is used to indicate the true elements in $\bar{y}$.
After a set of elements are processed, the corresponding elements
in $\bar{y}$ are updated with the true values from $y$ and $m$
is also updated accordingly. A deep CNN, implementing $h_{p}\left(\bar{y},m\right)$,
estimates the parameters of the Gaussian distribution for the elements
to be processed.

For the last scale, $y^{(s)}$, since there is no context, we use
the uni-variate non-parametric density model \cite{balle2018variational}
for the probability distribution. This model has been proved to be
effective in other LIC systems in modeling the hyperprior variables
and the implementations are readily available in many software packages
\cite{begaint2020compressai}.

Algorithm \ref{alg_pms} shows the operations of the MSP probability
model when estimating the rate loss for $y$ at the training stage.
Function $h_{s}\left(y^{(s)}\right)$ estimates the cross-entropy
of $y^{(s)}$ using the non-parametric density model and function
$h_{p}\left(\bar{y}^{(i)},m^{(i)}\right)$ outputs the estimated Gaussian
parameters for elements in $y^{(i)}$. The architecture of the deep
neural network implementing $h_{p}(\cdot)$ is shown in the appendix.

\begin{algorithm*}
\caption{Estimating the rate loss at the training stage \label{alg_pms}}

\begin{description}
\item [{given}] latent representation $y$ 
\item [{generate}] $n\sim U(-0.5,0.5)$ 
\item [{$y=y+n$}] ~ 
\item [{$\ensuremath{y^{(0)},y^{(1)},\cdots,y^{(s)}\leftarrow downsample(y)}\;\mathrm{(nearest\ neighbor\ downsampling)}$}]~
\item [{$r=-\log_{2}h_{s}\left(y^{(s)}\right)\ \mathrm{(rate\ loss\ of\ the\ last\ scale)}$}]~
\item [{for}] each $i$ in scale $s-1,\cdots,0$ 
\begin{description}
\item [{$\bar{y}^{(i)}=upsample\left(y^{(i+1)}\right)\ \mathrm{(nearest\ neighbor\ upsampling)}$}]~
\item [{$m^{(i)}=zero\left(y^{(i)}\right)$}]~
\item [{for}] each $j$ in subgroup $1,\cdots,b$ 
\begin{description}
\item [{for}] each $l$ in channel $1,\cdots,a$ 
\begin{description}
\item [{$\ensuremath{\mu_{(k,l)}^{(i,j)},\sigma_{(k,l)}^{(i,j)}\leftarrow h_{p}\left(\bar{y}^{(i)},m^{(i)}\right),}\ k=1,\cdots,d$}]~
\item [{$r=r+\left[-\sum_{k}\log_{2}p\left(y_{(k,l)}^{(i,j)}\vert\mu_{(k,l)}^{(i,j)},\sigma_{(k,l)}^{(i,j)}\right)\right]$}]~
\item [{$\bar{y}_{(k,l)}^{(i,j)}=y_{(k,l)}^{(i,j)},\ k=1,\cdots,d$}]~
\item [{$m_{(k,l)}^{(i,j)}=1,\ k=1,\cdots,d$}]~
\end{description}
\item [{$\ensuremath{r=r+\left[-\sum_{k}\log_{2}p\left(y_{(k,a+1,\cdots,c)}^{(i,j)}\vert\hat{y}^{(i)},m^{(i)}\right)\right]}$}]~
\end{description}
\end{description}
\item [{return}] $r$ 
\end{description}
\end{algorithm*}

\subsection{Latent representation overfitting}

At the inference stage, i.e. encoding and decoding, the latent representation
determines the size of the bitstream and the quality of the reconstructed
image. A LIC system can use the input image as the ground truth to
optimize the latent representation with the parameters of the decoder
and the probability model fixed \cite{campos2019content,zou2020l2c}.
This technique, known as latent representation overfitting (LOF),
can efficiently improve the compression performance by mitigating
the problem of domain shift \cite{krueger2021outofdistribution},
i.e., the characteristics of the input image significantly differing
from the images in the training dataset. The loss function of LOF
is defined as 
\begin{equation}
L=-\log p(y)+\lambda d\left(x,g_{s}\left(y\right)\right),\label{eq:overfitting_loss_fun}
\end{equation}
and optimized with regard to $y$. Note that directly optimizing Eq.
\ref{eq:overfitting_loss_fun} is difficult since $y$ is a discrete
tensor. Instead, we optimize the continuous latent representation
$\tilde{y}$ and quantize $\tilde{y}$ to get $y$. The objective
function can be written as 
\begin{equation}
L=-\log p\left(Q\left(\tilde{y}\right)\right)+\lambda d\left(x,g_{s}\left(Q\left(\tilde{y}\right)\right)\right),\label{eq:overfitting_loss_fun_q}
\end{equation}
where $Q\left(\cdot\right)$ is the quantization operation.

Since the gradient of the quantization function is undefined on border
values and zero elsewhere, a LIC system is normally trained by adding
uniform noise (AUN) to approximate the quantization operation at inference
time \cite{agustsson2020universally,balle2017endtoend}. However,
the AUN method requires a large number of samples and iterations to
converge. To optimize Eq. \ref{eq:overfitting_loss_fun_q} efficiently,
we opt for the straight-through estimation (STE) method \cite{bengio2013estimating}
to help the training during the LOF. Since the gradients are normally
small on a well-trained system, it may take many iterations to accumulate
the updates to change the values in $y$ using normal training parameters.
To solve this slow training problem, we apply a greedy search method
which is different from common training practice. We start the optimization
using a very large learning rate and record the best performing $\tilde{y}$
during the training. When the loss stops decreasing for a certain
number of iterations, we rewind $\tilde{y}$ to the best performing
state and continue the training with a lower learning rate. This procedure
repeats until no improvement can be made. Pseudocode for this procedure
is in the appendix.

\section{Experiments \label{sec:Experiments}}

\subsection{Profiles and training details}

The MSP probability model used in our LIC system contains a number
of parameters that affect the compression performance and the computational
complexity. We define three profiles for the proposed LIC system.
Table \ref{tab:profiles} shows the parameters for each profile and
the number of steps for the LIC system to decode an image.

\begin{table}
\caption{The profiles of the proposed LIC system used in the experiments. ``None''
in the channel seeds column means no channel dependency is exploited.
The filter column shows the number of filters of the deep CNN used
in the probability model. \label{tab:profiles}}

\smallskip{}

\centering{}%
\begin{tabular}{>{\raggedright}m{0.8cm}>{\centering}m{0.6cm}>{\centering}m{1.3cm}>{\centering}m{1cm}>{\centering}m{0.8cm}>{\centering}m{1cm}}
\toprule 
profile  & scales  & subgroup block  & channel seeds  & filters  & decoding steps\tabularnewline
\midrule 
baseline  & 3  & $2\times2$  & None  & 64  & 10\tabularnewline
normal  & 3  & $2\times2$  & 2  & 64  & 28\tabularnewline
extra  & 4  & $2\times4$  & 4  & 128  & 121\tabularnewline
\bottomrule
\end{tabular}

\vspace{-4mm}
\end{table}

The proposed LIC systems with different profiles are trained using
the Open Images dataset \cite{kuznetsova2020theopen}, which contains
images collected from the Internet. The majority of the images in
the Open Images dataset are compressed by JPEG \cite{jpeg}. The training
and validation splits of the dataset are prepared in the same way
as \cite{cao2020lossless,mentzer2019practical,mentzer2020learning,zhang2020lossless}.
The training split contains 340K images. The longest side of the training
images is 1024 pixels. For each profile, 7 models are trained with
different $\lambda$ values to achieve different compression rates.
The details of training and evaluation are in the appendix. 

\subsection{Kodak results}

Kodak dataset \cite{truecolor} is a popular dataset to benchmark
the performance of image compression systems. The dataset contains
24 uncompressed images with the size of $768\times512$ pixels. Table
\ref{tab:kodak_results} shows the compression performance of various
systems. In the table, ``vtm'' and ``hm'' are the reference implementations
of the VVC/H.266 and the HEVC/H.265 standards, respectively. The BD-rate
and BD-psnr are calculated using the VTM as the anchor. For the BD-rate,
a lower value indicates a better compression performance. For BD-psnr,
a higher value indicates a better performance. The numbers in the
table, except those for our proposed systems, are reported by the
CompressAI \cite{begaint2020compressai}. Note that encoding and decoding
are performed on CPU for VTM and HM while on GPU for the LIC systems.
Thus, the encoding and decoding time of the LIC systems are not directly
comparable to the VTM and HM systems. Following the protocol defined
in CompressAI, the encoding and decoding time of LIC systems do not
consider the model loading and data preparation time. RD curves of
the competing methods are in the appendix.

\begin{table*}[h]
\caption{Compression performance on the Kodak dataset.\label{tab:kodak_results}}

\smallskip{}

\centering{}%
\begin{tabular}{lccccc}
\toprule 
method  & BD-rate  & BD-psnr  & Parameters  & encoding time (s)  & decoding time (s)\tabularnewline
\midrule 
vtm \cite{bross2020versatile}  & 0.00  & 0.00  & -  & 125.87  & 0.13 \tabularnewline
hm \cite{sullivan2012overview}  & 23.31  & -1.01  & -  & 3.69  & 0.07 \tabularnewline
\midrule 
bmshj2018-hyperprior \cite{balle2018variational}  & 30.49  & -1.26  & 5.08M  & 0.04  & 0.03 \tabularnewline
mbt2018 \cite{minnen2018jointautoregressive}  & 11.02  & -0.49  & 14.13M  & 2.79  & 6.00 \tabularnewline
cheng2020-attn \cite{cheng2020learned}  & 5.51  & -0.23  & 13.18M  & 2.72  & 5.93 \tabularnewline
cheng2020-anchor \cite{cheng2020learned}  & 3.68  & -0.16  & 11.83M  & 2.75  & 5.94 \tabularnewline
\midrule 
ours: baseline  & 12.61  & -0.56  & 5.71M  & 0.23  & 0.20 \tabularnewline
ours: normal  & 11.12  & -0.50  & 5.79M  & 0.47  & 0.38 \tabularnewline
ours: extra  & 3.84  & -0.18  & 7.85M  & 1.43  & 1.18 \tabularnewline
\midrule 
ours: baseline (LOF)  & 4.35  & -0.20  & 5.71M  & 14.61  & 0.19 \tabularnewline
ours: normal (LOF)  & 3.44  & -0.16  & 5.79M  & 31.88  & 0.38 \tabularnewline
ours: extra (LOF)  & \textbf{-2.54 }  & \textbf{0.12 }  & 7.85M  & 122.01  & 1.41 \tabularnewline
\bottomrule
\end{tabular}
\end{table*}

As Table \ref{tab:kodak_results} shows, the proposed LIC system with
extra profile and LOF technique outperforms the competing systems,
including VTM. Compared to other LIC systems, the proposed systems
achieve similar or better performance with much smaller neural networks.
Furthermore, the proposed systems can decode images much more efficiently
than other LIC systems. Note that the difference between the baseline
and the normal profile is channel dependency exploitation. The relatively
small gain achieved by the normal profile indicates that the channels
in the latent representation are not heavily correlated, as we studied
in Section \ref{subsec:Channel-dependencies}.

Table \ref{tab:kodak_results} also shows significant gains that the
LOF technique brings by sacrificing the encoding time. However, this
is not a major issue for LIC systems since the encoding is normally
done offline and once. The encoding and decoding operations in the
traditional image and video codecs, such as VVC/H.266, are also asymmetric,
i.e., the encoding is much more complicated than the decoding.

\begin{figure*}
\begin{centering}
\begin{tabular}{cccccc}
 & uncompressed & VTM & Cheng2020-anchor & Extra & Extra (LOF)\tabularnewline
\multirow{2}{*}[1.3cm]{\includegraphics[width=5cm]{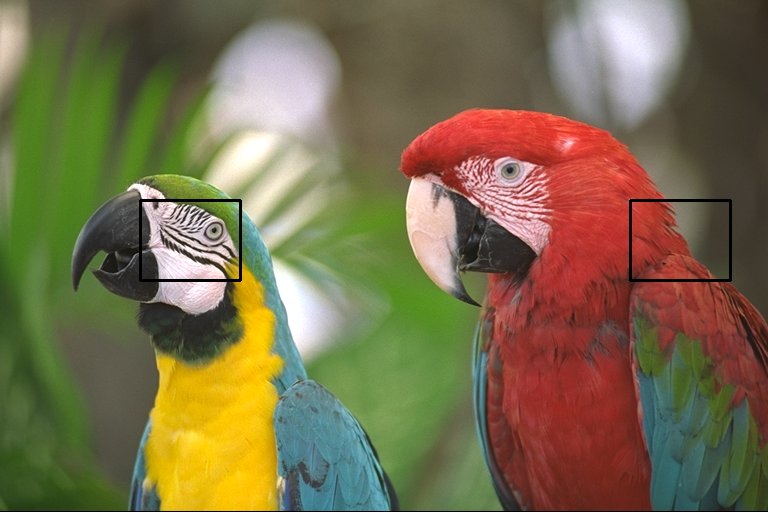}} & \includegraphics[width=2cm]{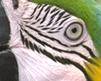} & \includegraphics[width=2cm]{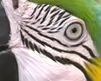} & \includegraphics[width=2cm]{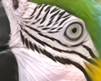} & \includegraphics[width=2cm]{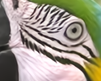} & \includegraphics[width=2cm]{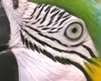}\tabularnewline
 & \includegraphics[width=2cm]{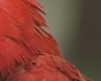} & \includegraphics[width=2cm]{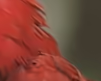} & \includegraphics[width=2cm]{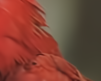} & \includegraphics[width=2cm]{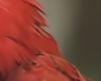} & \includegraphics[width=2cm]{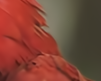}\tabularnewline
kodim23 &  & 0.238/37.37 & 0.235/37.59 & 0.229/37.25 & 0.220/37.57\tabularnewline
\multirow{2}{*}[1.3cm]{\includegraphics[width=5cm]{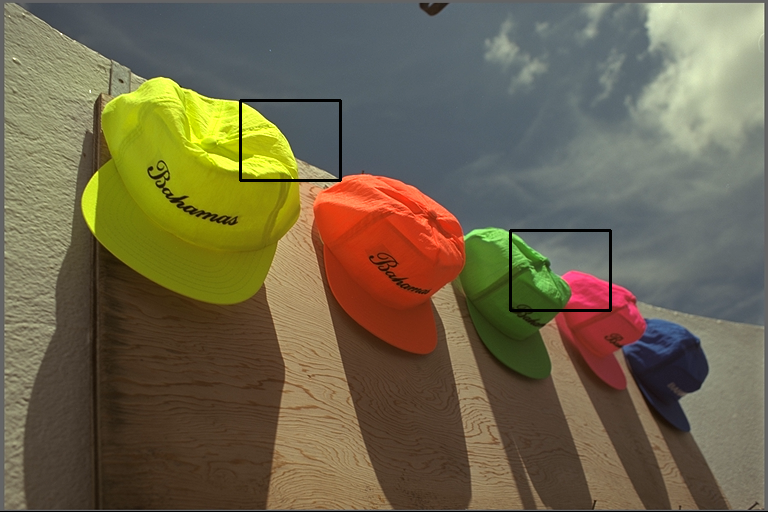}} & \includegraphics[width=2cm]{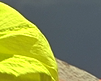} & \includegraphics[width=2cm]{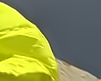} & \includegraphics[width=2cm]{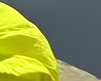} & \includegraphics[width=2cm]{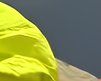} & \includegraphics[width=2cm]{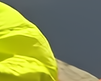}\tabularnewline
 & \includegraphics[width=2cm]{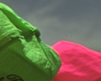} & \includegraphics[width=2cm]{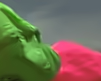} & \includegraphics[width=2cm]{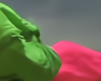} & \includegraphics[width=2cm]{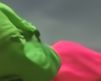} & \includegraphics[width=2cm]{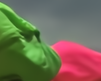}\tabularnewline
kodim03 &  & 0.114/33.72 & 0.128/33.94 & 0.118/33.82 & 0.109/33.96\tabularnewline
\multirow{2}{*}[1.3cm]{\includegraphics[width=5cm]{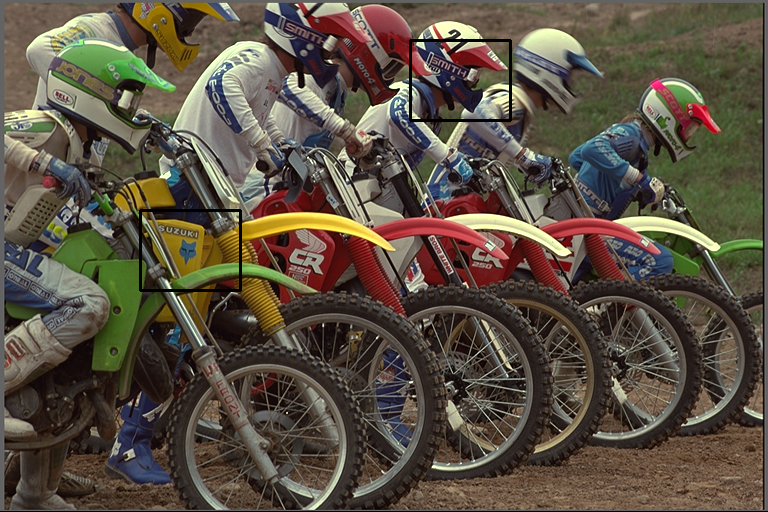}} & \includegraphics[width=2cm]{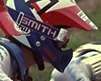} & \includegraphics[width=2cm]{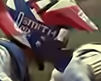} & \includegraphics[width=2cm]{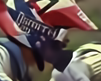} & \includegraphics[width=2cm]{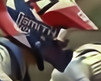} & \includegraphics[width=2cm]{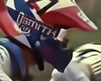}\tabularnewline
 & \includegraphics[width=2cm]{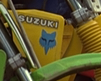} & \includegraphics[width=2cm]{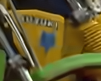} & \includegraphics[width=2cm]{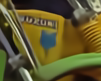} & \includegraphics[width=2cm]{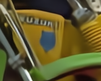} & \includegraphics[width=2cm]{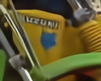}\tabularnewline
kodim05 &  & 0.239/25.94 & 0.240/26.09 & 0.243/26.16 & 0.232/26.28\tabularnewline
 &  &  &  &  & \tabularnewline
\end{tabular}
\par\end{centering}
\caption{Visualized comparison of selected methods. The numbers below the patches
are BPP/PSNR values. \label{fig:visualization_kodak} }

\vspace{-2mm}
\end{figure*}

Figure \ref{fig:visualization_kodak} visualizes some selected reconstructed
images from different systems on the Kodak dataset. For better visualization,
two patches from each image are shown in higher resolution. 

\subsection{CLIC2020 results}

Next, we evaluate the performance of the proposed LIC systems using
larger datasets. CLIC2020 dataset, divided into ``mobile'' and ``professional''
subsets, contains high-quality images. The average resolution of the
images in the ``mobile'' and ``professional'' subset is $1913\times1361$
and $1803\times1175$, respectively. The test splits of the two subsets
contain 178 and 250 images, respectively. Table \ref{tab:clic2020_results}
shows the compression performance of the competing systems.

\begin{table*}[h]
\caption{Compression performance on the``mobile'' and ``professional''
subset of the CLIC2020 dataset.\label{tab:clic2020_results}}

\smallskip{}

\centering{}%
\begin{tabular}{l|cc>{\centering}p{1.4cm}>{\centering}p{1.4cm}|cc>{\centering}p{1.4cm}>{\centering}p{1.4cm}}
\hline 
 & \multicolumn{4}{c|}{``mobile'' subset} & \multicolumn{4}{c}{``professional'' subset}\tabularnewline
\hline 
method  & BD-rate  & BD-psnr  & encoding time (s)  & decoding time (s)  & BD-rate  & BD-psnr  & encoding time (s)  & decoding time (s)\tabularnewline
\hline 
vtm \cite{bross2020versatile}  & 0.00  & 0.00  & 688.93  & 0.78  & 0.00  & 0.00  & 1970.66  & 1.29 \tabularnewline
hm \cite{sullivan2012overview}  & 21.86  & -0.96  & 24.63  & 0.46  & 26.56  & -0.99  & 47.86  & 0.67 \tabularnewline
\hline 
bmshj2018-hyperprior \cite{balle2018variational}  & 37.62  & -1.41  & 0.27  & 0.17  & 44.67  & -1.40  & 0.24  & 0.15 \tabularnewline
mbt2018 \cite{minnen2018jointautoregressive}  & 11.96  & -0.50  & 21.53  & 45.99  & 11.03  & -0.40  & 19.91  & 42.54 \tabularnewline
cheng2020-attn \cite{cheng2020learned}  & 8.92  & -0.35  & 21.10  & 45.15  & 5.97  & -0.21  & 19.94  & 43.07 \tabularnewline
cheng2020-anchor \cite{cheng2020learned}  & 8.29  & -0.32  & 21.26  & 45.92  & 4.08  & -0.14  & 19.84  & 42.76 \tabularnewline
\hline 
ours: baseline  & 15.74  & -0.69  & 0.68  & 0.64  & 21.27  & -0.80  & 0.69  & 0.68 \tabularnewline
ours: normal  & 14.69  & -0.64  & 0.92  & 0.80  & 20.92  & -0.79  & 0.97  & 0.86 \tabularnewline
ours: extra  & 8.02  & -0.36  & 2.76  & 2.20  & 11.72  & -0.46  & 2.54  & 2.04 \tabularnewline
\hline 
ours: baseline (LOF)  & 4.75  & -0.21  & 70.00  & 0.61  & 6.06  & -0.23  & 61.27  & 0.63 \tabularnewline
ours: normal (LOF)  & 4.12  & -0.18  & 104.76  & 0.82  & 5.30  & -0.20  & 102.29  & 0.85 \tabularnewline
ours: extra (LOF)  & \textbf{-1.04 }  & \textbf{0.04 }  & 423.46  & 2.18  & \textbf{-1.29 }  & \textbf{0.05 }  & 396.50  & 2.05 \tabularnewline
\hline 
\end{tabular}

\vspace{-4mm}
\end{table*}

Similar to the results on the Kodak dataset, the proposed LIC system
with extra profile and LOF technique outperforms other competing systems.
It is important to note that the LIC systems using the PixelCNN type
of context model are very inefficient in decoding large images. Compared
to the values in Table \ref{tab:kodak_results}, the decoding time
of these methods is proportional to the image size, which leads to
over 40 seconds to decode one image. With the help of the MSP probability
model, the decoding time of the proposed LIC system only increases
slightly with a constant number compared to the results on the Kodak
images. RD curves of the results on the CLIC2020 dataset are included
in the appendix. 

It can be noticed that the proposed methods perform better on the
Kodak dataset than on the CLIC2020 dataset. Furthermore, the performance
is better on the ``mobile'' subset than on the ``professional''
subset. This behavior may be caused by the training set we used to
train our models. First, the training images are about the same size
as the images in the Kodak dataset. Second, the quality of the training
images, intended for computer vision tasks and heavily compressed
using JPEG, are closer to the images in the ``mobile'' subset, while
significantly different from the high-quality images in the ``professional''
subset.

Using the LOF technique, the BD-rate increases by 6.38, 9.33 and 13.01
on the Kodak dataset, ``mobile'' subset and ``professional'' subset,
respectively. The significant gains demonstrate the effectiveness
of the LOF technique in mitigating the domain shift problem.

\vspace{-3mm}

\section{Discussions }

Recent research on LIC systems have shown comparable or superior performance
over the VVC/H.266 to some extent. For example, the Fu2021 \cite{fu2021learned}
system has a similar performance as the VVC/H.266 at the bit rate
less than 0.3 BPP and achieves 0.3 to 0.4 PSNR gain over the VVC/H.266
at the bit rate range {[}0.4, 0.9{]} BPP on the Kodak dataset. Ma2021
\cite{ma2021across} system claims a BD-rate reduction of 2.5\% at
bit range {[}0.1, 0.9{]} BPP on the Kodak dataset, and 2.2\% at bit
range {[}0.05, 0.5{]} BPP on the ``professional'' subset of the
CLIC2020 dataset. Guo2021 \cite{guo2021causalcontextual} LIC system
claims a BD-rate reduction of 5.1\% in bit rate range {[}0.15, 1{]}
BPP on the Kodak dataset.

Section \ref{sec:Experiments} shows that the proposed LIC system
achieves 2.54\%, 1.04\% and 1.29\% BD-rate reduction on three benchmark
datasets, respectively, compared to the VVC/H.266 over the bit rate
range {[}0.04, 2{]} BPP. To our best knowledge, the proposed system
is the first LIC system to achieve convincing BD-rate reduction over
the VVC/H.266 on the full bit-rate range on these benchmark datasets.

More importantly, the proposed LIC system has a much lower decoding
complexity compared to other LIC systems. The decoding complexity
of Fu2021 is equal to or higher than the Cheng2020 system, which takes
about 6 seconds to decode Kodak images and over 40 seconds to decode
CLIC2020 images \cite{cheng2020learned,fu2021learned}. Ma2021 system
claims that the decoding time is 3.8 times of the Cheng2020 system
\cite{ma2021across}. The Guo2021 system takes about 38.7 seconds
to decode Kodak images \cite{guo2021causalcontextual}. The decoding
complexity of the proposed system is determined by the configuration
constants. Experiments show that the proposed system is capable of
decoding images up to 2K at about 2 seconds. However, it should be
noted that the decoding time is also affected by the computational
capacity of the hardware system. It takes about 5.5 seconds to decode
a 4K image in our testing environment. We expect that this decoding
time on large images can be greatly reduced in an environment with
more computational capacity.

Experiment results also show that various balance points between compression
efficiency and computational complexity can be easily accomplished
by using predefined profiles. Some existing context models can be
considered as special cases of the proposed framework. For example,
the checkerboard context model defined in \cite{he2021checkerboard}
is the MSP profile with 1 scale downsampling (in a checkerboard manner)
and 1x1 subgroup block definition.

Although the LOF technique efficiently mitigates the domain-shift
problem that our system encounters because of the low quality training
images, we believe the performance of the proposed system can be further
improved if the system is trained with a large number of uncompressed
images.

We also trained the proposed system using MS-SSIM as the reconstruction
loss in Eq. \ref{eq:RD_loss}. The performance of the MS-SSIM trained
systems is demonstrated in the appendix. Further studies about the
impact of the profile parameters, such as scales $s$, subgroup size
$b$ and seed group size $a$ are also included in the appendix.

\vspace{-1mm}

\section{Conclusion}

In this paper, we proposed a novel LIC framework using an improved
MSP probability model and the latent representation overfitting technique.
The proposed system not only outperforms other LIC systems but also
the state-of-the-art codec VVC/H.266, which is based on traditional
technologies, on all three benchmark datasets over a wide bit rate
range. More importantly, the proposed system improves the decoding
complexity of previous LIC systems from $O(n)$ to $O(1)$, such that
the decoding time for 2K images are reduced by more than 20 times.
We also showed that the LOF technique efficiently mitigates the domain-shift
problem of a LIC system. The compression performance of the proposed
system can be further improved by various techniques and training
setups, which will be further studied in the future.

{\small{}

\bibliographystyle{IEEEtran}
\bibliography{clean}

\begin{thebibliography}{10}
\providecommand{\url}[1]{#1}
\csname url@samestyle\endcsname
\providecommand{\newblock}{\relax}
\providecommand{\bibinfo}[2]{#2}
\providecommand{\BIBentrySTDinterwordspacing}{\spaceskip=0pt\relax}
\providecommand{\BIBentryALTinterwordstretchfactor}{4}
\providecommand{\BIBentryALTinterwordspacing}{\spaceskip=\fontdimen2\font plus
\BIBentryALTinterwordstretchfactor\fontdimen3\font minus
  \fontdimen4\font\relax}
\providecommand{\BIBforeignlanguage}[2]{{%
\expandafter\ifx\csname l@#1\endcsname\relax
\typeout{** WARNING: IEEEtran.bst: No hyphenation pattern has been}%
\typeout{** loaded for the language `#1'. Using the pattern for}%
\typeout{** the default language instead.}%
\else
\language=\csname l@#1\endcsname
\fi
#2}}
\providecommand{\BIBdecl}{\relax}
\BIBdecl

\bibitem{jpeg}
JPEG, ``Jpeg - jpeg 2000,'' https://jpeg.org/jpeg2000/, 2022.

\bibitem{sullivan2012overview}
G.~J. Sullivan, J.-R. Ohm, W.-J. Han, and T.~Wiegand, ``Overview of the high
  efficiency video coding ({HEVC}) standard,'' \emph{IEEE Transactions on
  circuits and systems for video technology}, vol.~22, no.~12, pp. 1649--1668,
  2012, publisher: IEEE.

\bibitem{standardization2021isoiec2309032021}
I.~O.~f. Standardization, ``{ISO}/{IEC} 23090-3:2021 - {Information} technology
  {\textemdash} {Coded} representation of immersive media {\textemdash} {Part}
  3: {Versatile} video coding,'' 2021.

\bibitem{hu2021learning}
\BIBentryALTinterwordspacing
Y.~Hu, W.~Yang, Z.~Ma, and J.~Liu, ``Learning {End}-to-{End} {Lossy} {Image}
  {Compression}: {A} {Benchmark},'' \emph{arXiv:2002.03711 [cs, eess]}, Mar.
  2021, arXiv: 2002.03711. [Online]. Available:
  \url{http://arxiv.org/abs/2002.03711}
\BIBentrySTDinterwordspacing

\bibitem{marpe2006theh264mpeg4}
D.~Marpe, T.~Wiegand, and G.~Sullivan, ``The {H}.264/{MPEG4} advanced video
  coding standard and its applications,'' \emph{IEEE Communications Magazine},
  vol.~44, no.~8, pp. 134--143, Aug. 2006, conference Name: IEEE Communications
  Magazine.

\bibitem{tan2015videoquality}
T.~K. Tan, R.~Weerakkody, M.~Mrak, N.~Ramzan, V.~Baroncini, J.-R. Ohm, and
  G.~J. Sullivan, ``Video quality evaluation methodology and verification
  testing of {HEVC} compression performance,'' \emph{IEEE Transactions on
  Circuits and Systems for Video Technology}, vol.~26, no.~1, pp. 76--90, 2015,
  publisher: IEEE.

\bibitem{balle2017endtoend}
\BIBentryALTinterwordspacing
J.~Ball{\'e}, V.~Laparra, and E.~P. Simoncelli, ``End-to-end {Optimized}
  {Image} {Compression},'' \emph{arXiv:1611.01704 [cs, math]}, Mar. 2017,
  arXiv: 1611.01704. [Online]. Available: \url{http://arxiv.org/abs/1611.01704}
\BIBentrySTDinterwordspacing

\bibitem{cheng2020learned}
\BIBentryALTinterwordspacing
Z.~Cheng, H.~Sun, M.~Takeuchi, and J.~Katto, ``Learned {Image} {Compression}
  with {Discretized} {Gaussian} {Mixture} {Likelihoods} and {Attention}
  {Modules},'' \emph{arXiv:2001.01568 [eess]}, Mar. 2020, arXiv: 2001.01568.
  [Online]. Available: \url{http://arxiv.org/abs/2001.01568}
\BIBentrySTDinterwordspacing

\bibitem{guo2021causalcontextual}
\BIBentryALTinterwordspacing
Z.~Guo, Z.~Zhang, R.~Feng, and Z.~Chen, ``Causal {Contextual} {Prediction} for
  {Learned} {Image} {Compression},'' \emph{IEEE Transactions on Circuits and
  Systems for Video Technology}, pp. 1--1, 2021, arXiv: 2011.09704. [Online].
  Available: \url{http://arxiv.org/abs/2011.09704}
\BIBentrySTDinterwordspacing

\bibitem{hu2020coarsetofine}
Y.~Hu, W.~Yang, and J.~Liu, ``Coarse-to-{Fine} {Hyper}-{Prior} {Modeling} for
  {Learned} {Image} {Compression},'' in \emph{{AAAI} {Conference} on
  {Artificial} {Intelligenc}}, 2020.

\bibitem{minnen2018jointautoregressive}
\BIBentryALTinterwordspacing
D.~Minnen, J.~Ball{\'e}, and G.~Toderici, ``Joint {Autoregressive} and
  {Hierarchical} {Priors} for {Learned} {Image} {Compression},''
  \emph{arXiv:1809.02736 [cs]}, Sep. 2018, arXiv: 1809.02736. [Online].
  Available: \url{http://arxiv.org/abs/1809.02736}
\BIBentrySTDinterwordspacing

\bibitem{theis2017lossyimage}
\BIBentryALTinterwordspacing
L.~Theis, W.~Shi, A.~Cunningham, and F.~Husz{\'a}r, ``Lossy {Image}
  {Compression} with {Compressive} {Autoencoders},'' \emph{arXiv:1703.00395
  [cs, stat]}, Mar. 2017, arXiv: 1703.00395. [Online]. Available:
  \url{http://arxiv.org/abs/1703.00395}
\BIBentrySTDinterwordspacing

\bibitem{choi2019variable}
\BIBentryALTinterwordspacing
Y.~Choi, M.~El-Khamy, and J.~Lee, ``Variable {Rate} {Deep} {Image}
  {Compression} {With} a {Conditional} {Autoencoder},'' \emph{arXiv:1909.04802
  [cs, eess]}, Sep. 2019, arXiv: 1909.04802. [Online]. Available:
  \url{http://arxiv.org/abs/1909.04802}
\BIBentrySTDinterwordspacing

\bibitem{patel2019humanperceptual}
\BIBentryALTinterwordspacing
Y.~Patel, S.~Appalaraju, and R.~Manmatha, ``Human {Perceptual} {Evaluations}
  for {Image} {Compression},'' \emph{arXiv:1908.04187 [cs, eess]}, Aug. 2019,
  arXiv: 1908.04187. [Online]. Available: \url{http://arxiv.org/abs/1908.04187}
\BIBentrySTDinterwordspacing

\bibitem{mentzer2019conditional2}
\BIBentryALTinterwordspacing
F.~Mentzer, E.~Agustsson, M.~Tschannen, R.~Timofte, and L.~Van~Gool,
  ``Conditional {Probability} {Models} for {Deep} {Image} {Compression},''
  \emph{arXiv:1801.04260 [cs]}, Jun. 2019, arXiv: 1801.04260. [Online].
  Available: \url{http://arxiv.org/abs/1801.04260}
\BIBentrySTDinterwordspacing

\bibitem{zhao2021auniversal2}
J.~Zhao, B.~Li, J.~Li, R.~Xiong, and Y.~Lu, ``A {Universal} {Encoder} {Rate}
  {Distortion} {Optimization} {Framework} for {Learned} {Compression},'' in
  \emph{2021 {IEEE}/{CVF} {Conference} on {Computer} {Vision} and {Pattern}
  {Recognition} {Workshops} ({CVPRW})}, Jun. 2021, pp. 1880--1884, iSSN:
  2160-7516.

\bibitem{he2021checkerboard}
\BIBentryALTinterwordspacing
D.~He, Y.~Zheng, B.~Sun, Y.~Wang, and H.~Qin,
  ``\BIBforeignlanguage{en}{Checkerboard {Context} {Model} for {Efficient}
  {Learned} {Image} {Compression}},'' in \emph{\BIBforeignlanguage{en}{2021
  {IEEE}/{CVF} {Conference} on {Computer} {Vision} and {Pattern} {Recognition}
  ({CVPR})}}, Jun. 2021. [Online]. Available:
  \url{https://ieeexplore.ieee.org/document/9577406/}
\BIBentrySTDinterwordspacing

\bibitem{balle2018variational}
\BIBentryALTinterwordspacing
J.~Ball{\'e}, D.~Minnen, S.~Singh, S.~J. Hwang, and N.~Johnston, ``Variational
  image compression with a scale hyperprior,'' \emph{arXiv:1802.01436 [cs,
  eess, math]}, May 2018, arXiv: 1802.01436. [Online]. Available:
  \url{http://arxiv.org/abs/1802.01436}
\BIBentrySTDinterwordspacing

\bibitem{begaint2020compressai}
J.~B{\'e}gaint, F.~Racap{\'e}, S.~Feltman, and A.~Pushparaja, ``{CompressAI}: a
  {PyTorch} library and evaluation platform for end-to-end compression
  research,'' \emph{arXiv preprint arXiv:2011.03029}, 2020.

\bibitem{fu2021learned}
\BIBentryALTinterwordspacing
H.~Fu, F.~Liang, J.~Lin, B.~Li, M.~Akbari, J.~Liang, G.~Zhang, D.~Liu, C.~Tu,
  and J.~Han, ``Learned {Image} {Compression} with {Discretized}
  {Gaussian}-{Laplacian}-{Logistic} {Mixture} {Model} and {Concatenated}
  {Residual} {Modules},'' \emph{arXiv:2107.06463 [cs, eess]}, Jul. 2021, arXiv:
  2107.06463. [Online]. Available: \url{http://arxiv.org/abs/2107.06463}
\BIBentrySTDinterwordspacing

\bibitem{gao2021neuralimage}
G.~Gao, P.~You, R.~Pan, S.~Han, Y.~Zhang, Y.~Dai, and H.~Lee, ``Neural {Image}
  {Compression} via {Attentional} {Multi}-{Scale} {Back} {Projection} and
  {Frequency} {Decomposition},'' in \emph{Proceedings of the {IEEE}/{CVF}
  {International} {Conference} on {Computer} {Vision} ({ICCV})}, Oct. 2021, pp.
  14\,677--14\,686.

\bibitem{ma2021across}
\BIBentryALTinterwordspacing
C.~Ma, Z.~Wang, R.~Liao, and Y.~Ye, ``A {Cross} {Channel} {Context} {Model} for
  {Latents} in {Deep} {Image} {Compression},'' \emph{arXiv:2103.02884 [cs,
  eess]}, Mar. 2021, arXiv: 2103.02884. [Online]. Available:
  \url{http://arxiv.org/abs/2103.02884}
\BIBentrySTDinterwordspacing

\bibitem{xie2021enhanced}
Y.~Xie, K.~L. Cheng, and Q.~Chen, ``Enhanced invertible encoding for learned
  image compression,'' \emph{arXiv preprint arXiv:2108.03690}, 2021.

\bibitem{he2021checkerboard2}
\BIBentryALTinterwordspacing
D.~He, Y.~Zheng, B.~Sun, Y.~Wang, and H.~Qin, ``Checkerboard {Context} {Model}
  for {Efficient} {Learned} {Image} {Compression},'' \emph{arXiv:2103.15306
  [cs, eess]}, Apr. 2021, arXiv: 2103.15306. [Online]. Available:
  \url{http://arxiv.org/abs/2103.15306}
\BIBentrySTDinterwordspacing

\bibitem{wang2003multiscale}
Z.~Wang, E.~Simoncelli, and A.~Bovik, ``Multiscale structural similarity for
  image quality assessment,'' in \emph{The {Thrity}-{Seventh} {Asilomar}
  {Conference} on {Signals}, {Systems} {Computers}, 2003}, vol.~2, Nov. 2003,
  pp. 1398--1402 Vol.2.

\bibitem{chen2019neuralimage}
\BIBentryALTinterwordspacing
T.~Chen, H.~Liu, Z.~Ma, Q.~Shen, X.~Cao, and Y.~Wang, ``Neural {Image}
  {Compression} via {Non}-{Local} {Attention} {Optimization} and {Improved}
  {Context} {Modeling},'' \emph{arXiv:1910.06244 [eess]}, Oct. 2019, arXiv:
  1910.06244. [Online]. Available: \url{http://arxiv.org/abs/1910.06244}
\BIBentrySTDinterwordspacing

\bibitem{minnen2020channelwise}
\BIBentryALTinterwordspacing
D.~Minnen and S.~Singh, ``Channel-wise {Autoregressive} {Entropy} {Models} for
  {Learned} {Image} {Compression},'' \emph{arXiv:2007.08739 [cs, eess, math]},
  Jul. 2020, arXiv: 2007.08739. [Online]. Available:
  \url{http://arxiv.org/abs/2007.08739}
\BIBentrySTDinterwordspacing

\bibitem{Jain_Abbeel_Pathak_2020}
\BIBentryALTinterwordspacing
A.~Jain, P.~Abbeel, and D.~Pathak, ``Locally masked convolution for
  autoregressive models,'' \emph{arXiv:2006.12486 [cs, stat]}, Jun 2020.
  [Online]. Available: \url{http://arxiv.org/abs/2006.12486}
\BIBentrySTDinterwordspacing

\bibitem{oord2016conditional}
\BIBentryALTinterwordspacing
A.~v.~d. Oord, N.~Kalchbrenner, O.~Vinyals, L.~Espeholt, A.~Graves, and
  K.~Kavukcuoglu, ``Conditional {Image} {Generation} with {PixelCNN}
  {Decoders},'' \emph{arXiv:1606.05328 [cs]}, Jun. 2016, arXiv: 1606.05328.
  [Online]. Available: \url{http://arxiv.org/abs/1606.05328}
\BIBentrySTDinterwordspacing

\bibitem{salimans2016pixelcnn}
\BIBentryALTinterwordspacing
T.~Salimans, A.~Karpathy, X.~Chen, and D.~P. Kingma, ``{PixelCNN}++:
  {Improving} the {PixelCNN} with {Discretized} {Logistic} {Mixture}
  {Likelihood} and {Other} {Modifications},'' Nov. 2016. [Online]. Available:
  \url{https://openreview.net/forum?id=BJrFC6ceg}
\BIBentrySTDinterwordspacing

\bibitem{cao2020lossless}
\BIBentryALTinterwordspacing
S.~Cao, C.-Y. Wu, and P.~Kr{\"a}henb{\"u}hl, ``\BIBforeignlanguage{en}{Lossless
  {Image} {Compression} through {Super}-{Resolution}},''
  \emph{\BIBforeignlanguage{en}{arXiv:2004.02872 [cs, eess]}}, Apr. 2020,
  arXiv: 2004.02872. [Online]. Available: \url{http://arxiv.org/abs/2004.02872}
\BIBentrySTDinterwordspacing

\bibitem{zhang2020lossless}
H.~Zhang, F.~Cricri, H.~R. Tavakoli, N.~Zou, E.~Aksu, and M.~M. Hannuksela,
  ``Lossless {Image} {Compression} {Using} a {Multi}-{Scale} {Progressive}
  {Statistical} {Model},'' in \emph{Proceedings of the {Asian} {Conference} on
  {Computer} {Vision} ({ACCV})}, Nov. 2020.

\bibitem{truecolor}
\BIBentryALTinterwordspacing
Kodak, ``True {Color} {Kodak} {Images},'' 2022. [Online]. Available:
  \url{http://r0k.us/graphics/kodak/}
\BIBentrySTDinterwordspacing

\bibitem{campos2019content}
\BIBentryALTinterwordspacing
J.~Campos, S.~Meierhans, A.~Djelouah, and C.~Schroers, ``Content {Adaptive}
  {Optimization} for {Neural} {Image} {Compression},'' \emph{arXiv:1906.01223
  [cs, eess]}, Jun. 2019, arXiv: 1906.01223. [Online]. Available:
  \url{http://arxiv.org/abs/1906.01223}
\BIBentrySTDinterwordspacing

\bibitem{zou2020l2c}
\BIBentryALTinterwordspacing
N.~Zou, H.~Zhang, F.~Cricri, H.~R. Tavakoli, J.~Lainema, M.~Hannuksela,
  E.~Aksu, and E.~Rahtu, ``{L2C} -- {Learning} to {Learn} to {Compress},''
  \emph{Proceedings of the IEEE 22nd International Workshop on Multimedia
  Signal Processing (MMSP)}, Jul. 2020, arXiv: 2007.16054. [Online]. Available:
  \url{http://arxiv.org/abs/2007.16054}
\BIBentrySTDinterwordspacing

\bibitem{krueger2021outofdistribution}
\BIBentryALTinterwordspacing
D.~Krueger, E.~Caballero, J.-H. Jacobsen, A.~Zhang, J.~Binas, D.~Zhang, R.~L.
  Priol, and A.~Courville, ``Out-of-{Distribution} {Generalization} via {Risk}
  {Extrapolation} ({REx}),'' \emph{arXiv:2003.00688 [cs, stat]}, Feb. 2021,
  arXiv: 2003.00688. [Online]. Available: \url{http://arxiv.org/abs/2003.00688}
\BIBentrySTDinterwordspacing

\bibitem{agustsson2020universally}
\BIBentryALTinterwordspacing
E.~Agustsson and L.~Theis, ``Universally {Quantized} {Neural} {Compression},''
  \emph{arXiv:2006.09952 [cs, math, stat]}, Oct. 2020, arXiv: 2006.09952.
  [Online]. Available: \url{http://arxiv.org/abs/2006.09952}
\BIBentrySTDinterwordspacing

\bibitem{bengio2013estimating}
\BIBentryALTinterwordspacing
Y.~Bengio, N.~L{\'e}onard, and A.~Courville, ``Estimating or {Propagating}
  {Gradients} {Through} {Stochastic} {Neurons} for {Conditional}
  {Computation},'' \emph{arXiv:1308.3432 [cs]}, Aug. 2013, arXiv: 1308.3432.
  [Online]. Available: \url{http://arxiv.org/abs/1308.3432}
\BIBentrySTDinterwordspacing

\bibitem{kuznetsova2020theopen}
A.~Kuznetsova, H.~Rom, N.~Alldrin, J.~Uijlings, I.~Krasin, J.~Pont-Tuset,
  S.~Kamali, S.~Popov, M.~Malloci, A.~Kolesnikov, T.~Duerig, and V.~Ferrari,
  ``The {Open} {Images} {Dataset} {V4}: {Unified} image classification, object
  detection, and visual relationship detection at scale,'' \emph{IJCV}, 2020.

\bibitem{mentzer2019practical}
\BIBentryALTinterwordspacing
F.~Mentzer, E.~Agustsson, M.~Tschannen, R.~Timofte, and L.~Van~Gool,
  ``Practical {Full} {Resolution} {Learned} {Lossless} {Image} {Compression},''
  \emph{arXiv:1811.12817 [cs, eess]}, May 2019, arXiv: 1811.12817. [Online].
  Available: \url{http://arxiv.org/abs/1811.12817}
\BIBentrySTDinterwordspacing

\bibitem{mentzer2020learning}
\BIBentryALTinterwordspacing
F.~Mentzer, L.~Van~Gool, and M.~Tschannen, ``Learning {Better} {Lossless}
  {Compression} {Using} {Lossy} {Compression},'' \emph{arXiv:2003.10184 [cs,
  eess]}, Mar. 2020, arXiv: 2003.10184. [Online]. Available:
  \url{http://arxiv.org/abs/2003.10184}
\BIBentrySTDinterwordspacing

\bibitem{bross2020versatile}
B.~Bross, J.~Chen, S.~Liu, and Y.-K. Wang, ``Versatile {Video} {Coding}
  ({Draft} 8),'' \emph{Joint Video Experts Team (JVET), Document JVET-Q2001},
  Jan. 2020.

\end{thebibliography}

%
%
%

\end{document}